\documentclass[letterpaper, 10 pt, journal, twoside]{IEEEtran}  % Comment this line out if you need a4paper
\usepackage{amsmath,amsfonts}
\usepackage{algorithmic}
\usepackage{algorithm}
\usepackage{array}
\usepackage[caption=false,font=normalsize,labelfont=sf,textfont=sf]{subfig}
\usepackage{textcomp}
\usepackage{stfloats}
\usepackage{url}
\usepackage{verbatim}
\usepackage{graphicx}
\usepackage{cite}
\usepackage{tabularx}
\usepackage{threeparttable}
\usepackage{multirow} 
\usepackage{color}
\usepackage{gensymb}
\usepackage[section]{placeins}
\usepackage{float}
\usepackage{makecell}
\usepackage[switch]{lineno}
\usepackage{booktabs}
\usepackage[table]{xcolor}
\usepackage{pifont}

\begin{document}
% \linenumbers

\title{Residual Learning towards High-fidelity Vehicle Dynamics Modeling with Transformer}

\author{
Jinyu Miao$^{1}$, Rujun Yan$^{1}$, Bowei Zhang$^{1}$, Tuopu Wen$^{1,*}$, \\Kun Jiang$^{1,*}$, Mengmeng Yang$^{1}$, Jin Huang$^{1}$, Zhihua Zhong$^{2}$, Diange Yang$^{1,*}$ 
% <-this % stops a space
\thanks{This work was supported in part by the National Natural Science Foundation of China (52394264, 52472449, 52372414, 52402499), Beijing Natural Science Foundation (23L10038, L231008), Beijing Municipal Science and Technology Commission (Z241100003524013, Z241100003524009), and China Postdoctoral Science Foundation (2024M761636).}% <-this % stops a space
\thanks{$^{1}$Jinyu Miao, Rujun Yan, Bowei Zhang, Tuopu Wen, Kun Jiang, Mengmeng Yang, Jin Huang, and Diange Yang are with the School of Vehicle and Mobility, Tsinghua University, Beijing, China.{\tt\small jinyu.miao97@gmail.com}}%
\thanks{$^{2}$Zhihua Zhong is with Chinese Academy of Engineering, Beijing, China.}%
\thanks{$^{*}$Corresponding author: Diange Yang, Kun Jiang, and Tuopu Wen}
}

\markboth{Journal of \LaTeX\ Class Files,~Vol.~14, No.~8, August~2021}%
{Shell \MakeLowercase{\textit{et al.}}: A Sample Article Using IEEEtran.cls for IEEE Journals}

% \newcommand{\benny}[1]{\textcolor{red}{[#1]}}

\maketitle

\begin{abstract}
    The vehicle dynamics model serves as a vital component of autonomous driving systems, as it describes the temporal changes in vehicle state. In a long period, researchers have made significant endeavors to accurately model vehicle dynamics.
    Traditional physics-based methods employ mathematical formulae to model vehicle dynamics, but they are unable to adequately describe complex vehicle systems due to the simplifications they entail. 
    Recent advancements in deep learning-based methods have addressed this limitation by directly regressing vehicle dynamics. However, the performance and generalization capabilities still require further enhancement.
    In this letter, we address these problems by proposing a vehicle dynamics correction system that leverages deep neural networks to correct the state residuals of a physical model instead of directly estimating the states. 
    This system greatly reduces the difficulty of network learning and thus improves the estimation accuracy of vehicle dynamics. 
    Furthermore, we have developed a novel Transformer-based dynamics residual correction network, DyTR. This network implicitly represents state residuals as high-dimensional queries, and iteratively updates the estimated residuals by interacting with dynamics state features.
    The experiments in simulations demonstrate the proposed system works much better than physics model, and our proposed DyTR model achieves the best performances on dynamics state residual correction task, reducing the state prediction errors of a simple 3 DoF vehicle model by an average of 92.3\% and 59.9\% in two dataset, respectively.
\end{abstract}


\section{Introduction}
\label{sec:intro}

\IEEEPARstart{A}{cquiring} high-precision models of complex vehicle dynamical systems is critical in Autonomous Driving (AD) researches, due to its ability to truthfully mimic a real vehicle's dynamics, thus helping to develop and tune vehicle planning and control modules \cite{Akar2014,lian2015cornering}.
Inaccurate state predictions from the dynamic models will result in an incorrectly optimized trajectory and eventually lead to inappropriate vehicle motions, hindering the safety of AD systems \cite{li2024ego}.

\begin{figure}[!t]
    \centering
    \includegraphics[width=0.97\linewidth]{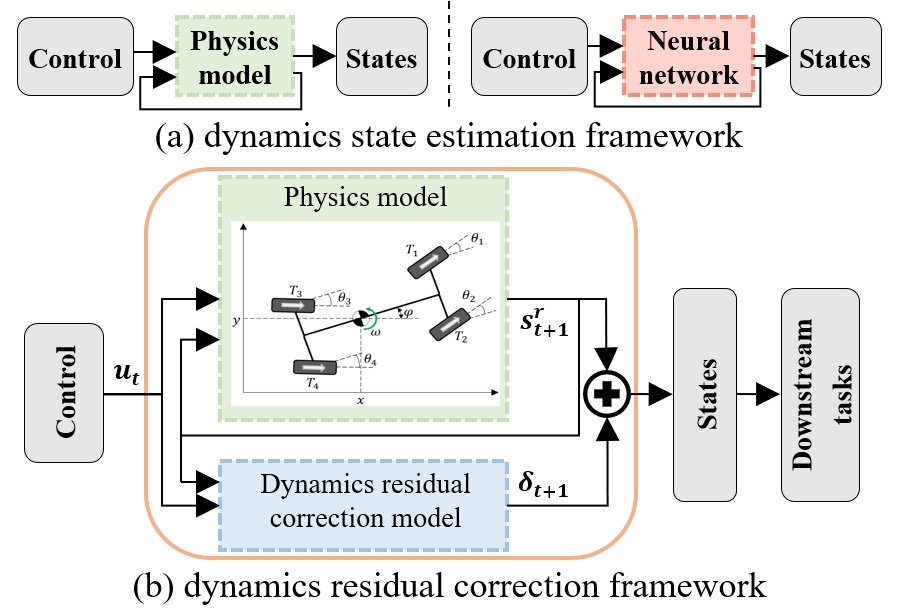}
    \caption{The diagram of (a) the Dynamic State Estimation (DSE) methods and (b) the Dynamics Residual Correction (DRC) methods.}
    \label{fig:intro}
\end{figure}

Over the decades of development in vehicle dynamics research, the field has witnessed significant advancements and paradigm shifts. 
Some researches attempt to directly approximate vehicle dynamics, which is categorized as the Dynamics State Estimation (DSE) framework in this letter, as shown in Fig. \ref{fig:intro} (a).
Early solutions under the DSE framework focus on developing sophisticated physics-based models, from simple point-mass model \cite{Kapania}, single-track model \cite{Timings2013}, to double-track model \cite{Perantoni04052014}.
This evolution signifies the deepening of researchers' understanding of complex vehicle systems, and has led to a gradual increase in the precision and Degree-of-Freedom (DoF) of vehicle dynamics state estimation.
However, these physics-based vehicle dynamics models inevitably drift for long-horizon state prediction due to simplification in modeling. This phenomenon is even significant when the DoF of vehicle is increased (\textit{e.g.}, distributed electric-driven vehicles) or the operating conditions of vehicles are challenging. 
Inspired by recent significant successes taken by deep learning techniques in several areas \cite{resnet, silver2016mastering}, researchers have begun to employ Deep Neural Networks (DNN) model for directly learning vehicle dynamical attributes. 
Being attributable to the exceptional non-linear fitting capabilities of DNN models, it achieves superior performance on vehicle dynamic modeling task than physics-based methods and gets rid of the limited human knowledge of vehicle systems \cite{spielberg2019neural,Xu2019}. Such a solution learns DNN model in data-driven manner, so its accuracy and generalization ability may be challenged due to varying distributions of dynamics states between different conditions. And the performance of long-range state estimation is still limited due to its open-loop training strategy. 

As researchers have found in the ResNet \cite{resnet}, \textit{the DNN models are generally easier to learn residual functions with reference than directly learn unreferenced functions, and can gain performance improvements from residual learning}. Therefore, the residual learning scheme is widely adopted in recent proposed DNN models \cite{vaswani2017attention,tolstikhin2021mlp}. Motivated by such a notion, researchers propose the Dynamics Residual Correction (DRC) framework that employ existing advanced NN models to learn dynamics state residuals of a dynamics based model instead of directly learning dynamics state \cite{DRC-baidu,DRC-baidu2}, as shown in Fig. \ref{fig:intro} (b). Since the dynamics state residuals are commonly caused by simplifications of based model and highly non-linear corrected factors of complex vehicle systems, their value ranges and distributions are more similar under different conditions than dynamics states, so these DRC methods perform better than DSE methods with a lot of margins \cite{DRC-baidu, DRC-baidu2}.

However, current models applied in the DRC framework are general networks adopted from other research area, such as MLP and Transformer \cite{vaswani2017attention}, which do not take into account the specificity of the DRC task, so there is room for improvement in their performance by designing specific DRC networks. 

Furthermore, electric-drive vehicles are gradually replacing the dominance of internal combustion engine drive vehicles as AD technology develops. Some leading AD companies develop distributed electric-drive vehicles, \textit{e.g.}, Tesla Model S. But the related DNN-based methods towards complex distributed electric-drive vehicles modeling are limited, blocking the development of AD researches in this area.

With these context in mind, in this letter, we propose Dynamic residuals correction Transformer (DyTR) model, which learns to predict residuals of a based model (a 3 DoF physics-based model in this letter) conditioned on the the based model's estimation and implicit vehicle dynamics through supervised learning. Specifically, we encode vehicle dynamics by historical states and control signals. The estimation of based model and vehicle configuration are embedded as \textit{dynamics query}. A Transformer-based state estimation module is then adopted to associate \textit{dynamics query} and vehicle dynamics features. Finally, the updated \textit{dynamics query} can be decoded into state residuals.
Benefit from the proper network structure, our proposed DyTR network could significantly reduce the state estimation errors of the based model. The primary contributions are summarized as:

\begin{itemize}
    \item To the best knowledge of the authors, DyTR is the first work that adopts DRC framework into distributed electric-drive vehicles, allowing high-fidelity vehicle dynamics analysis without effort-costly and time-consuming physics modeling and parameter identification.
    
    \item We propose a Transformer-based model, DyTR, which is a specific DRC network that could effectively learn to estimate dynamic state residuals of the base model. The proper network structure makes DyTR work better than general DNN models.

    \item We generate corresponding vehicle dynamics dataset for distributed electric-drive vehicles via co-simulation. Through comprehensive experiments, we show the validity of our proposed model, which achieve the state-of-the-art performance on DRC task. We also prove the rationality of model setups by detailed ablation analysis. 
\end{itemize}

\section{Related Works}
\label{sec:review}

Modeling vehicle dynamics has been studied for decades. Here, we classify and review related work into two folders as in Sec. \ref{sec:intro}, namely, DSE methods and DRC methods.

\subsection{Vehicle Dynamics State Estimation (DSE) methods}

The DSE methods can be further spitted into three classes, \textit{i.e.}, pure physics-based ones, pure DNN-based ones, and the hybrid ones. Each of them has achieved remarkable success in the last few decades. 

Physics-based DSE methods, or analytic methods, model vehicle dynamics as mathematical rules, and they can vary greatly in complexity and precision. 
As the most simplistic model, point-mass model \cite{Kapania} neglects the size of vehicles and load transfer caused by lateral and longitudinal acceleration, and only considers the vehicle as an infinitesimally small point with a mass, which is commonly used for online trajectory generation \cite{Subosits2019}. 
Single-track models, or 3 DoF bicycle models, \cite{Timings2013} improve realism by considering the geometry of the vehicle, providing a more accurate representation of the vehicle’s steering. They project the front and rear wheels on two virtual wheels located along the longitudinal axis of the vehicle and estimate the forces applied to different components of the vehicle. Single-track approaches can capture three fundamental DoF of a vehicle, so they are widely used in planning and control \cite{Liu2018}. Later, to capture additional effects like yaw moments from different drive and brake forces across an axle, scholars develop double-track models \cite{Perantoni04052014}. 
% In \cite{Berntorp03102014}, the authors found a little high-level variation in optimal trajectories generated with single-track and double-track models, while \cite{Gottmann2018} highlighted the importance of longitudinal load transfer by comparing point-mass and single-track models. 
In \cite{Subosits2021}, the authors quantified the impacts of model fidelity on the effectiveness of trajectory optimization and found that the dynamics model used does not have a major effect on the optimal path or speed for the vehicle, but simple models fail to capture transient dynamics.
The trade-off problem between accuracy and efficiency remains an unsolved problem for physics modeling.
Furthermore, System identification is necessary for physics-based DSE models.
Current solutions widely adopt Gaussian process \cite{Kabzan2019} or NN methods \cite{COSTA2023104469} to identify parameters of the dynamics systems, but it is still difficult to capture the overall dynamics of the complex vehicle systems due to the limited dimensions of the state-space systems \cite{Johan2019}, and the identification process requires significant time and commercial costs of real vehicle calibrations. 

Inspired by recent achievements in many research fields \cite{resnet,vaswani2017attention,poet}, AD researchers and engineers have developed DNN-based DSE approach to directly regress vehicle dynamical systems, instead of deriving traditional analytic formula. 
As early as 2015, Punjani \textit{et al.} represented the helicopter dynamics with a Rectified Linear Unit (ReLU) Network Model \cite{Punjani2015}.
In \cite{spielberg2019neural}, Spielberg \textit{et al.} proposed to employ a DNN to predict yaw and lateral acceleration given a sequence of past states and control inputs, and also experimentally demonstrated that the DNN achieved better performance than the physical model when implemented in the same feedforward-feedback control architecture. 
As an early exploration by Baidu Apollo team, \cite{Xu2019} evaluated several novel DNN architectures in vehicle dynamics regression and demonstrated the DNN-based models can achieve higher accuracy with significantly reduced re-development effort compared to rule-based ones. 
Hermansdorfer \textit{et al.} investigated the performance of Gated Recurrent Unit (GRU) networks in DSE task \cite{Hermansdorfer2020}.
Cao \textit{et al.} further improved by designing two DNN models to separately learn longitudinal and lateral dynamical properties of the vehicle \cite{Cao2021}. 
Since the neural network generally contains millions of parameters and high non-linearity, it has demonstrated significant potential to learn to faithfully describe the complex and non-linear dynamics of vehicles from large volumes of data, thus achieving improved state estimation accuracy then physics-based DSE methods, but its accuracy and generalization ability are challenged limited due to domain shift between different conditions.

In recent five years, aiming at high precision and deployable vehicle dynamics modeling, researchers have explored hybrid DSE methods combining physics-based dynamics models with advanced DNN models, sometimes called Physics-Informed Neural Network (PINN) in this area. 
For instance, in \cite{Kim2022}, Kim \textit{et al.} proposed a full differentiable physics model with prior knowledge by the Pacejka tire model \cite{Pacejka01011992} and 3DoF bicycle model. later, Chrosniak \textit{et al.} improved by ensuring that the internal coefficient estimates remain within their nominal physical ranges \cite{Chrosniak2024}. Similar to system identification approaches in physics-based methods, these two methods also estimated physics coefficient and then leverage well-established dynamical equations to accurately predict vehicle states, but the overall processes are achieved by neural networks in end-to-end manner. These hybrid DSE methods achieve higher precision than traditional pure physics-based or DNN-based alternates, but its upper-bound is still limited by the involved physical dynamics model.

\subsection{Vehicle Dynamics Residual Correction (DRC) methods}

In theory, the DRC methods are also hybrid methods, but they combine physics-based dynamics models with advanced DNN models in another strategy that the DNN model commonly takes physics model's estimation as input and regress the errors (or residuals) of physics model, instead of directly regress dynamics states.
The Baidu Apollo team firstly proposed the DRC framework \cite{DRC-baidu,DRC-baidu2} in 2021. The framework is composed of two models: 1) an open-loop physics-based vehicle dynamics model, and 2) a DNN model that uses Stochastic Variational Gaussian Process (SVGP) to model estimation residuals of physics model.
Ning \textit{et al.} also used Gaussian Process (GP) to model dynamics residual, but they further combined deep kernel learning with GP for DRC task \cite{pmlr-v229-ning23a}.
Chen \textit{et al.} focused on lateral velocity estimation tasks and utilized transfer learning scheme to broaden the applicability of DRC methods, allowing fast model transfer across different vehicle classes \cite{Chen2024}. 
Baier \textit{et al.} employ DRC framework for interpretable vehicle state estimation, restricting the output range of the DNN model as part of the hybrid system, which limits the uncertainty of the DNN model \cite{Baier2021HybridPA}.
These methods all explore the ability of DNN model to regress residuals to refine the estimation of base methods, thus pushing the limitation of physics-based dynamics model.
 
Our work also employ the same DRC framework, but we put our attention on designing a proper DNN model to achieve significant improvements, filling the gap in the DRC approach to specific network designing, and we work towards more complex distributed electric-driven vehicles.

\section{Methodology}
\label{sec:method}

In this section, we first format the DRC problem. Then, a data collection pipeline through the co-simulation by MATLAB and CarSim software is presented. Finally, we introduce the proposed Transformer-based DRC network in details. 

\subsection{Problem Statement}
\label{sec:problem}

As a well-known state estimation problem, vehicle dynamics modeling has been studied for more than a hundred years. In principle, an idealistic vehicle dynamics model $\texttt{H}(\cdot)$ can predict future vehicle states $\textbf{s}_{t+1}$ given current states $\textbf{s}_{t}$, current control signals $\textbf{u}_{t}$, and vehicle configuration $c$: $\textbf{s}_{t+1}=\texttt{H}(\textbf{s}_{t},\textbf{u}_{t},c)$, where $t$ is the timestamp.

In this work, we employ the DRC framework in which a traditional 3 DoF bicycle model serves as the physics-based state estimator, called \textit{base model}, and a DNN model acts to correct the estimated vehicle states, as shown in Fig. \ref{fig:intro} (b). Specifically, the \textit{base model} $\texttt{F}_{base}(\cdot)$ takes $\hat{\textbf{s}}_{t}$, $\textbf{u}_{t}$, $c$, and then estimates vehicle states in the future timestamp:

\begin{equation}
\label{equ:1}
    \hat{\textbf{s}}_{t+1}=\texttt{F}_{base}(\hat{\textbf{s}}_{t},\textbf{u}_{t},c)
\end{equation}
where $\hat{(\cdot)}$ means the estimated states by the \textit{base model}. 
Due to unavoidable simplification and limited human knowledge about vehicle dynamics, $\hat{\textbf{s}}$ often have large estimation errors, especially in long-term close-loop estimation. 
Therefore, the dynamics residuals can be formatted as the difference between real vehicle dynamics states and the estimated states:
\begin{equation}
\label{equ:residual}
    \delta=\textbf{s}-\hat{\textbf{s}}
\end{equation}

To predicts the residuals and refine the estimation of the \textit{base model}, the DNN model is fed by a sequence of $T$-step historical estimated vehicle states $\mathcal{S}={\{\hat{\textbf{s}}_{t-T+1},..., \hat{\textbf{s}}_{t}\}}$, a sequence of $T$-step historical control signals $\mathcal{U}={\{\textbf{u}_{t-T+1},..., \textbf{u}_{t}\}}$, $c$, and $\hat{\textbf{s}}_{t+1}$, and generates an estimation of residuals $\hat{\delta}_{t+1}$:

\begin{equation}
\label{equ:2}
    \hat{\delta}_{t+1}=\texttt{F}_{nn}(\mathcal{S},\mathcal{U},c,\hat{\textbf{s}}_{t+1})
\end{equation}
Finally, the vehicle states are corrected as $\hat{\textbf{s}}_{t+1} + \hat{{\delta}}_{t+1}$.

It should be noticed that the \textit{based model} runs in closed-loop, independent of DNN model. And the \textit{based model} can be any existing physics-based \cite{Timings2013,Perantoni04052014} or DNN-based \cite{spielberg2019neural,Kim2022,Pacejka01011992} vehicle dynamics models. This work chooses 3 DoF model to validate the effectiveness for its simplicity.

\subsection{Vehicle State Data Collection}
Due to the lack of publicly available dynamics states data for distributed electric-drive vehicles, we generate required data through the simulation. The detailed pipeline is outlined below.

\label{sec:data}

{
\setlength{\parindent}{0cm}
\textbf{Base Model}
First, we develop the commonly used 3 DoF and 14 DoF vehicle models to distributed electric-driven fashion in MATLAB Simulink software, where the torque and steering angle of the wheels can be controlled independently. 
For both models, we use the driving torque $T_{1} \sim T_{4}$ and the steering angles $\theta_{1} \sim \theta_{4}$ of four wheels as the control signals, and focus on three vehicle dynamics state, namely, linear longitudinal velocity $v_x$, linear latitudinal velocity $v_y$, and the acceleration of yaw angle $w_z$. Therefore, the control signals $ \textbf{u}$ and vehicle dynamics states \textbf{s} studied in this work are defined as:

\begin{equation}
    \textbf{u} = [T_{1} \sim T_{4} ~ \theta_{1}\sim \theta_{4}],~\textbf{s}=[v_x, v_y,w_z]
\end{equation} 
}

\begin{figure}[!t]
    \centering
    \includegraphics[width=0.97\linewidth]{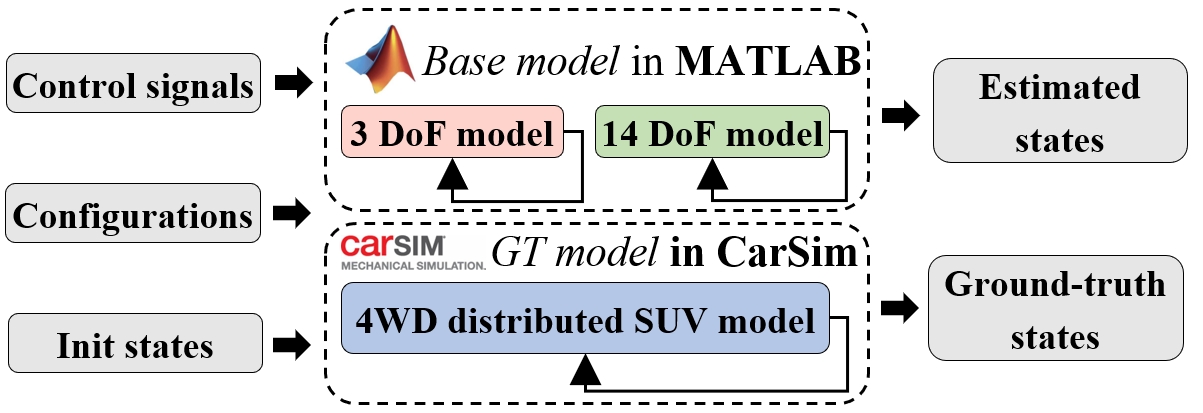}
    \caption{The diagram of data generation pipeline through co-simulation by MATLAB and CarSim.}
    \label{fig:data}
\end{figure}

{
\setlength{\parindent}{0cm}
\textbf{Data Generation by Co-Simulation}
Following common practices in AD researches, we employ the distributed electric-driven vehicle in CarSim simulator, as \textit{Ground-Truth (GT) model}, to collect GT vehicle dynamics states data.
We firstly align the basic vehicle configuration parameters between the \textit{GT model} in the CarSim and \textit{base models} in the MATLAB, such as vehicle dimensions, vehicle masses, and the dynamics of the wheel-ground interaction, as shown in Fig. \ref{fig:data}. During the co-simulation process, the same vehicle control signals are simultaneously applied to both the \textit{base models} and the \textit{GT model}. All the models independently run in closed-loop, so each simulation only requires an initial vehicle states and the temporal control signals, which are also aligned between the \textit{base models} and the \textit{GT model}.
}

Next, we generate proper control signals for distributed electric-driven vehicles. The values for steering angles and driving torques are modeled using sine waves, allowing us to simulate various driving conditions, including acceleration, deceleration, steady-speed driving, and steering. To ensure vehicle stability and rationality of conditions, we maintain an opposite relationship between the front and rear wheels' steering angles throughout the simulation and manually verify all the generated data. 
Furthermore, to prove the adaptation ability of the DRC method towards vehicles with varying configuration, we also collect state data with different vehicle mass. So, the vehicle configuration $c$ considered in this work is set to the vehicle mass.

At each simulation step, the vehicle states generated by the \textit{base models} in MATLAB are collected as estimated states, $\hat{\textbf{s}}$, and the ones from CarSim \textit{GT model} are collected as GT vehicle states, $\textbf{s}$. 
Finally, we obtain a vehicle dynamics state dataset that contains approximately 761 thousands valid data and split the collected dataset into three part as described in Tab. \ref{tab:data}. The \textbf{val1} split aims to evaluate the performances of DRC methods with various vehicle configuration while the \textbf{val2} split provides a more challenging benchmark that evaluates the performance with different conditions and vehicle.

% seq_len
\begin{table}[t]
	\centering
	\caption{The detailed description of three splits in the generated dataset.}
    \renewcommand\arraystretch{1.5}
	\begin{threeparttable}
    \resizebox{\linewidth}{!}{
		\begin{tabular}{r | r | c }
			\toprule
			split & num. & description \\
			\midrule
			\textbf{train} & 626472 & 28 driving conditions and 9 vehicle mass\\
            \hline
            \textbf{val1} & 67122 & \makecell{28 driving conditions (same as \textbf{train}) \\ and a vehicle mass (different from \textbf{train})} \\
            \hline
            \textbf{val2} & 67122 & \makecell{28 driving conditions (different from \textbf{train}) \\ and a vehicle mass (different from \textbf{train})} \\
			\bottomrule
		\end{tabular}
    }
	\end{threeparttable}
	\label{tab:data}
\end{table}

\subsection{Transformer-based Dynamics Residual Correction}
\label{sec:network}

{
\setlength{\parindent}{0cm}
\textbf{Preliminaries: Transformer attention \cite{vaswani2017attention}}
We apply attention mechanism from Transformer in this work. For better readability, we briefly review attention as background. As the key component in Transformer-based works \cite{carion2020end,poet}, attention layers take $d$-dimensional query vector $q$, key vector $k$, and value vector $v$ and conduct a following calculation:
}
\begin{equation}
\label{equ:attn}
    \texttt{Attention}(q,k,v)=\texttt{softmax}(\frac{qk^T}{\sqrt{d}})v
\end{equation}
Intuitively, $q$ retrieves related information from $v$ based on the relevance between $q$ and $k$. Then, the retrieved values are used to update $q$ in the Transformer layer. 

This core insight behind this work is inspired by residual learning \cite{resnet} that the residuals in state estimation are mainly caused by the simplification of the \textit{base model}, which are generally highly nonlinear but low-amplitude, making it easier for the network learn to estimate residuals compared to estimating states. 
Also, the residuals are related to two components, \textit{i.e.}, vehicle dynamics and the estimation of the \textit{base model}. Therefore, we first learn vehicle dynamics from historical states $\mathcal{S}$ and control signals $\mathcal{U}$, and then estimate the residuals by further considering the estimation of the \textit{base model} $\hat{\textbf{s}}_{t+1}$, that is, the Eqn. \ref{equ:2} can be reformulated as :
\begin{equation}
\label{equ:4}
    \hat{\delta}_{t+1}=\texttt{F}_{nn}(\hat{\textbf{s}}_{t+1},c~|~\mathcal{S},\mathcal{U})
\end{equation}

\begin{figure*}[!t]
    \centering
    \includegraphics[width=0.97\linewidth]{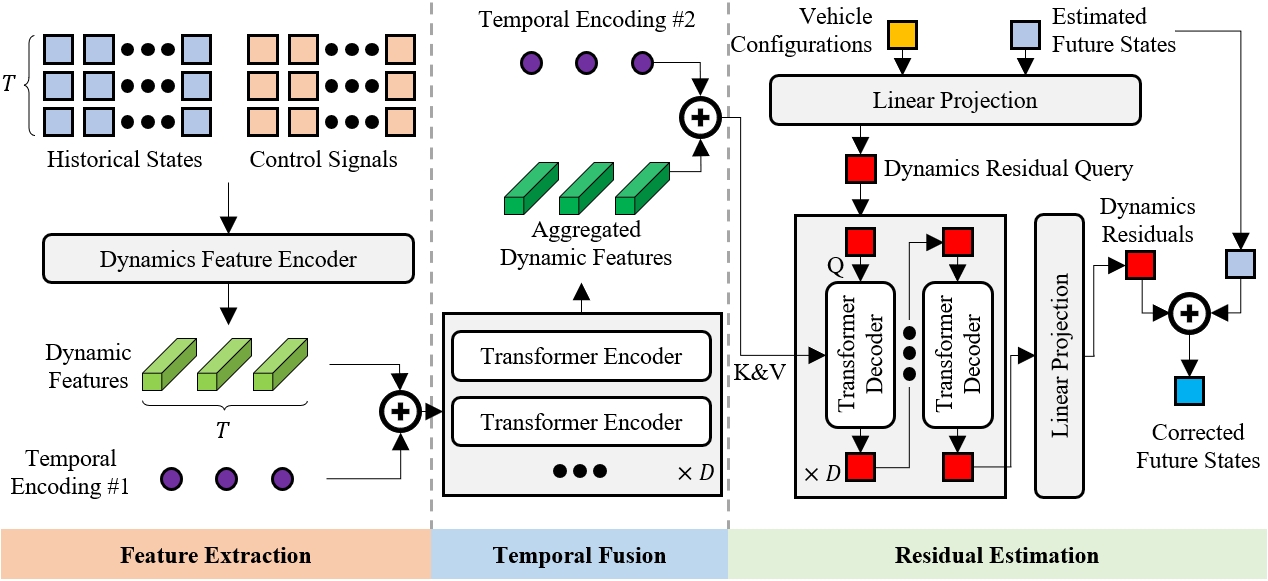}
    \caption{The network structure of our proposed Transformer-based DRC model, DyTR. The model takes historical $T$-step states, $T$-step control signals, vehicle configurations, and estimated future states by the \textit{base model} as input, and estimates the residuals of dynamics states.}
    \label{fig:network}
\end{figure*}

Following successful attempts in DNN-based dynamic modeling methods \cite{spielberg2019neural,Xu2019}, we build a network composed of three modules, including feature extraction, temporal fusion, and residual estimation, as shown in Fig. \ref{fig:network}. 

{
\setlength{\parindent}{0cm}
\textbf{Feature Extraction} First, we individually concatenate the vehicle control signals and estimated states in each timestamp and project them into high-dimensional dynamics feature $E^s_i$:
}

\begin{equation}
\label{equ:5}
    E^s_i = \texttt{F}_{enc}([{\textbf{s}}^r_{i}, \textbf{u}_{i}]),i \in [t-N+1,t] 
\end{equation}
where $\texttt{F}_{enc}(\cdot)$ is the dynamics feature encoder that can be any DNN model, \textit{e.g.}, MLP and Kolmogorov–Arnold Networks (KAN) \cite{liu2024kan}, and $[\cdot, \cdot]$ denotes channel-wise concatenation.

{
\setlength{\parindent}{0cm}
\textbf{Temporal Fusion} In related vehicle dynamics modeling works, researchers believed that the high-dimensional dynamics features after temporal fusion contain vehicle dynamics characteristics \cite{spielberg2019neural,Xu2019}. Therefore, after extracting dynamic features from each timestamp, we utilize a Transformer-based module to fuse temporal dynamics features. We utilize well-established temporal Transformer encoder modules to internally interact dynamics features along the temporal dimension and get aggregated dynamic feature $E^f$:
}

\begin{equation}
\label{equ:6}
    E^f = \mbox{Encoder}(E^s,\mbox{pos}=P^s)
\end{equation}
It should be noticed that encoder module discards the temporal order during calculation. A dynamic feature should attract different attention in different temporal positions. Therefore, we further add a learnable temporal embedding $P^s$ to the dynamics features to maintain their temporal order information.

{
\setlength{\parindent}{0cm}
\textbf{Residual Estimation}
Different from previous works that directly estimate dynamic states from fused dynamic feature \cite{spielberg2019neural,Xu2019,Hermansdorfer2020,Cao2021}, we instead estimate dynamics state residuals. Inspired by recent success by query-based works with Transformer \cite{carion2020end, vaswani2017attention, poet}, We also format the DRC task in a query-based scheme. Concretely, we project the estimated future states of the \textit{base model} $\hat{\textbf{s}}_{t+1}$and the vehicle configuration $c$ to high-dimensional feature $Q$, called dynamics residual query, by linear projection:
}

\begin{equation}
\label{equ:7}
    Q = \texttt{Linear}([{\textbf{s}}^r_{t+1}, c])
\end{equation}
Then we utilize Transformer decoder modules to iteratively update query feature based on retrieved information conditioned on aggregated dynamic features:

\begin{equation}
\label{equ:8}
    Q = \mbox{Decoder}(q=Q,k=v=E^f,\mbox{pos}=P^q)
\end{equation}
where $P^q$ is another learnable temporal embedding. In the decoder module, implicit relevant information can be estimated by feature relationship between $Q$ and $E^f$, and be used to update $Q$ so that $Q$ gradually contains dynamics state residual information by feature-level interactions. Finally, we decode $Q$ to dynamic state residuals by linear projection:

\begin{equation}
\label{equ:9}
    \hat{{\delta}}_{t+1} = \texttt{Linear}(Q)
\end{equation}
The estimated vehicle states can be corrected by $\hat{\textbf{s}}_{t+1}+\hat{\delta}_{t+1}$.

To train the parameters of the DRC network, we employ weighted smooth L1 loss between corrected states and ground-truth states from the \textit{GT model}:

\begin{equation}
\label{equ:10}
    \mathcal{L} = \sum_{i \in [v_x, v_y, w_z]} \alpha_i \cdot {||(\hat{\delta}_{t+1}(i)+\hat{\textbf{s}}_{t+1}(i))-\textbf{s}_{t+1}(i)||}_{S1}
\end{equation}
where $\alpha$ denotes the weighting coefficients used to balance different vehicle states.

\section{Experiments}
\label{sec:result}

\begin{table*}[!t]
	\centering
	\caption{The accuracy of vehicle dynamic state estimation of different physics-based DSE methods and DNN-based DCR methods.}
    \renewcommand\arraystretch{1.5}
	\begin{threeparttable}
    \resizebox{\linewidth}{!}{
		\begin{tabular}{r | r | r r r | r r r | r r r | r r r }
			\toprule
			\multirow{2}{*}{model} & \multirow{2}{*}{param.} & \multicolumn{3}{c | }{Mean error on \textbf{val1} $\downarrow$ } & \multicolumn{3}{c|}{Max error on \textbf{val1} $\downarrow$} & \multicolumn{3}{c|}{Mean error on \textbf{val2} $\downarrow$}  & \multicolumn{3}{c}{Max error on \textbf{val2} $\downarrow$} \\
             & & $v_x$ & $v_y$ & $w_z$ & $v_x$ & $v_y$ & $w_z$  & $v_x$  & $v_y$ & $w_z$  & $v_x$ & $v_y$ & $w_z$  \\
			\midrule
            \multicolumn{14}{l}{\textbf{Physics-based DSE methods}} \\
            \hline
			3 DoF model & - & 0.19 & 0.035 & 0.0080 & 0.34 & 0.15 & 0.023 & 0.13 & 0.015 & 0.0042 & 0.23 & 0.028 & 0.010  \\
            14 DoF model & - & 0.16 & 0.029 & 0.0081 & 0.29 & 0.13 & 0.023 & 0.12 & 0.014 & 0.0044 & 0.22 & 0.051 & 0.010 \\
            \hline
            % \multicolumn{1}{l|}{3 DoF model + } \\
            \multicolumn{14}{l}{\textbf{DNN-based DRC methods}} \\
            \hline
            \multicolumn{1}{r|}{MLP(64d)} & 116.8K & 0.10 & 0.020 & 0.0063 & 0.31 & 0.09 & 0.022 & 0.082 & 0.0088 & 0.0029 & 0.20 & 0.037 & 0.0089  \\
            \multicolumn{1}{r|}{KAN(64d)} & 707.9K & 0.090 & 0.039 & 0.0048 & 0.28 & 0.12 & 0.021 & 0.10 & 0.013 & 0.0040 & 0.23 & 0.045 & 0.012 \\
            \multicolumn{1}{r|}{MLP+Trans(64d)} & 102.5K & 0.046 & 0.0086 & 0.0019 & 0.16 & 0.039 & 0.011 & 0.12 & 0.011 & 0.0026 & 0.29 & 0.034 & 0.0093 \\
            \multicolumn{1}{r|}{KAN+Trans(64d)} & 127.1K & 0.059 & 0.017 & 0.0057 & 0.22 & 0.086 & 0.023 & 0.094 & 0.0096 & 0.0029 & 0.23 & 0.037 & 0.0095 \\
            \multicolumn{1}{r|}{Informer(64d) \cite{zhou2021informer}} & 351.4K & 0.078 & 0.020 & 0.0064 & 0.26 & 0.079 & 0.020 & 0.099 & 0.0091 & 0.0036 & 0.23 & 0.029 & 0.0088 \\
            \rowcolor{gray!20} \multicolumn{1}{r|}{KANDyTR(64)} & 218.0K & 0.025 & \textbf{0.0041} & \textbf{0.00072} & 0.11 & \textbf{0.022} & \textbf{0.0049} & 0.074 & 0.0094 & 0.0022 & 0.28 & 0.076 & 0.013 \\
            \rowcolor{gray!20} \multicolumn{1}{r|}{MLPDyTR(64)} & 170.8K & \textbf{0.022} & 0.0044 & 0.00079 & \textbf{0.10} & 0.024 & 0.0053 & \textbf{0.059} & \textbf{0.0062} & \textbf{0.0013} & \textbf{0.19} & \textbf{0.022} & \textbf{0.0059} \\
            \hline
             \rowcolor{gray!40} \multicolumn{1}{r|}{MLPDyTR(128d)} & 537.6K & \textbf{0.016} & \textbf{0.0027} & \textbf{0.00057} & \textbf{0.080} & \textbf{0.021} & \textbf{0.0046} & 0.071 & \textbf{0.0052} & \textbf{0.0013} & \textbf{0.19} & \textbf{0.022} & 0.0069 \\
             \rowcolor{gray!40} & & 91.6\%$\downarrow$  & 92.3\%$\downarrow$ & 92.9\%$\downarrow$ & 76.5\%$\downarrow$ & 86.0\%$\downarrow$ & 80.0\%$\downarrow$ & 45.4\%$\downarrow$ & 65.3\% $\downarrow$ & 69.0\%$\downarrow$ & 17.4\%$\downarrow$ & 21.4\%$\downarrow$ & 31.0\%$\downarrow$ \\
			\bottomrule
		\end{tabular}
    }
    \begin{tablenotes}
        \footnotesize
        \item[] All the DCR methods employ 3 DoF model as the \textit{base model}.
    \end{tablenotes}
	\end{threeparttable}
	\label{table:main}
\end{table*}

\subsection{Experimental Settings}
{
\setlength{\parindent}{0cm}
\textbf{Implementation Details}
The 3 DoF and 14 DoF models are simulated by MATLAB2023b/Simulink software, we choose 3 DoF model as the \textit{base model} in this work. The \textit{GT model} is developed from the full size 4WD distributed SUV from CarSim 2020.0 simulator.
The proposed DRC network is implemented using PyTorch library and trained from scratch for 100 epoches using the ADAM optimizer with a batch size of 256, an initial learning rate of $1e^{-3}$, and weight decay of $1e^{-5}$ on a single GeForce RTX 3090 GPU. The weighting coefficients in loss function (Eqn. \ref{equ:10}) is set as: $\alpha_{v_x} = 1, \alpha_{v_y}=10, \alpha_{w_z}=1000$. By default, we use MLP in feature extraction module and set the feature dimension $C$ as 64, the length of temporal sequence $T$ is 15, and the depth of encoder and decoder layers $D$ are both set as 2 for balance between performance and efficiency.  
}

{
\setlength{\parindent}{0cm}
\textbf{Compared Baselines}
To prove the effectiveness of the proposed DyTR model in DRC task, we adopt some DNN models with the same three-module network structure, and feed $\hat{\textbf{s}}_{t+1}$ and $c$ into the feature extraction module so that the inputs are the same as the proposed DyTR model for fair comparison. The details are described below:
}

\begin{itemize}
    \item \textbf{MLP} only utilizes MLP modules to build a simplest network. After extracting $E^s_i$ by a MLP module for each timestamp, the temporal $T$ features are flatten and aggregated by another MLP module for temporal fusion. The residuals are estimated by the third MLP module.
    \item \textbf{KAN} replaces all the MLP modules in \textbf{MLP} with the recently proposed KAN module \cite{liu2024kan}.
    \item \textbf{MLP+Trans.} employs transformer-based temporal fusion module to aggregate temporal $T$ features. 
    \item \textbf{KAN+Trans.} replaces MLP-based feature extraction module in \textbf{MLP+Trans.} with the KAN module \cite{liu2024kan}.
    \item \textbf{Informer} \cite{zhou2021informer} is an efficient Transformer-based time-series forecasting network. We directly regulate its input format to make it suitable for the DRC task.
\end{itemize}
We set the feature dimension, the length of temporal sequence, the depth of transformer layers of all the compared methods as the same as default configuration of our DyTR model.

\subsection{Main Results}

\begin{figure}[!t]
    \centering
    \includegraphics[width=0.97\linewidth]{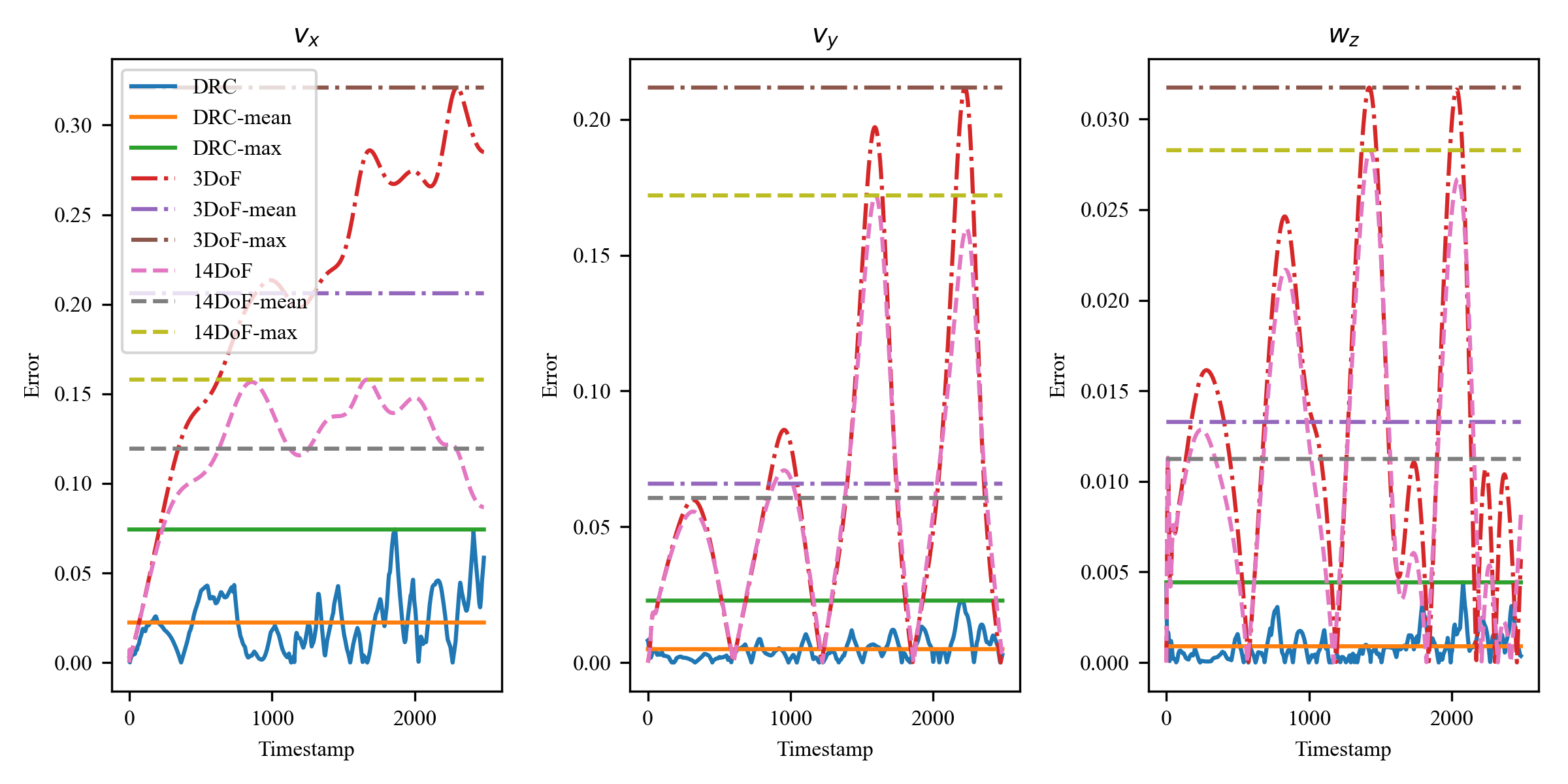}
    \caption{A case of the state correction performance by DyTR on the \textbf{val1} split.}
    \label{fig:exp}
\end{figure}

Tab. \ref{table:main} provides the performances of our proposed DyTR model and compared baselines in DRC task. 
On the both \textbf{val1} and \textbf{val2} split, the 14 DoF model generally exhibits higher state estimation accuracy than the 3 DoF model, indicating that simplifications and linearizations in physical-based vehicle modeling can lead to a degradation in model fidelity. 
However, the more intricate 14 DoF model imposes more strict requirements on parameter identification, and its performance improvement over lower DoF models is relatively modest.
When employing DNN models to correct the residuals of state predictions from the \textit{base model}, the hybrid models demonstrate a significant improvement in the accuracy of state estimation. 
Even the simple \textbf{MLP} model can substantially reduce the errors of state estimation. 
The proposed DyTR model achieves notably effective performance whether using MLP or KAN \cite{liu2024kan} as the feature extraction module, \textit{i.e.}, MLPDyTR and KANDyTR. Fig. \ref{fig:exp} shows a case of the state correction by our proposed MLPDyTR on the \textbf{val1} split. Compared to other DNN-based baselines, the DyTR model is more effective and efficient in reducing the state estimation errors of the \textit{base model}.

By comparing the performance of MLPDyTR with \textbf{MLP+Trans.}, and KANDyTR with \textbf{KAN+Trans.}, it clearly proves the effectiveness of the proposed Transformer-based state residuals estimation module. The Transformer is capable of iteratively extracting relevant implicit information from the aggregated dynamic features $E^f$ and using it to update the dynamics residual query $Q$, enabling the DyTR models to progressively approximate the state residuals of the \textit{base model} with high precision.

Equally noteworthy is the fact that our model demonstrates commendable performances across both the \textbf{val1} and \textbf{val2} splits. The efficacy on \textbf{val1} substantiates the generalization capability of the DyTR model with respect to vehicle configurations, while its performance on \textbf{val2} underscores its adeptness at generalizing across diverse driving conditions.

Finally, we also evaluate a MLPDyTR variant with large feature dimension $C=128$. The experimental results reveal that as the feature dimensions increase, the model's performance further improves. On \textbf{val1} split, the model is able to reduce the average state estimation error of the \textit{base model} by an average of 92.3\%. On more challenging \textbf{val2} split, it also achieves a 59.9\% reduction in average error, demonstrating the model's outstanding potential in the DRC task.

\subsection{Ablation Analysis}

To better understand the effectiveness of each modules and select the best network configuration of our proposed DyTR model, we conduct detailed ablation study to validate through a series of experiments. 

% seq_len
\begin{table}[t]
	\centering
	\caption{The performance of vehicle dynamics state correction with different temporal length.}
    \renewcommand\arraystretch{1.5}
	\begin{threeparttable}
    \resizebox{\linewidth}{!}{
		\begin{tabular}{r | r r r | r r r }
			\toprule
			\multirow{2}{*}{$T$} & \multicolumn{3}{c | }{Mean error on \textbf{val1} $\downarrow$ } & \multicolumn{3}{c}{Mean error on \textbf{val2} $\downarrow$}\\
             & $v_x$ & $v_y$ & $w_z$ & $v_x$ & $v_y$ & $w_z$  \\
			\midrule
			1 & 0.038 & 0.0092 & 0.0023 & 0.11 & 0.024 & 0.0092 \\
            5 & 0.051 & 0.012 & 0.0023 & 0.052 & 0.0081 & 0.0019\\
            10 & 0.028 & 0.0053 & 0.00087 & 0.060 & 0.0070 & 0.0015\\
            \rowcolor{gray!40}15 & \textbf{0.022} & 0.0044 & \textbf{0.00079} & \textbf{0.059} & \textbf{0.0062} & \textbf{0.0013} \\
            20 & \textbf{0.022} & 0.0043 & 0.00093 & 0.095 & 0.0068 & 0.0022\\
            25 & 0.024 & \textbf{0.0042} & 0.00088 & 0.067 & 0.0097 & 0.0027\\
            30 & 0.031 & 0.0049 & 0.0011 & 0.16 & 0.0098 & 0.0028\\
			\bottomrule
		\end{tabular}
    }
    \begin{tablenotes}
        \footnotesize
        \item[] All the DCR methods employ MLPDyTR model.
    \end{tablenotes}
	\end{threeparttable}
	\label{table:seq_len}
\end{table}

\subsubsection{Effectiveness of different temporal length:} To investigate the effects of different temporal state length $T$, we test the performances of MLPDyTR model with different $T$. The results are listed in Tab. \ref{table:seq_len}. It can be observed that as $T$ increases, the state correction efficacy initially improves progressively. However, once $T$ exceeds 15, the state correction performance begins to decline. This indicates that an appropriate length of historical states and control signals data can help model learn vehicle dynamics modeling. Conversely, an excessive volume of data can introduce noise and disrupt the model learning. Therefore, we select $T=15$ as the default configuration.

% depth
\begin{table}[t]
	\centering
	\caption{The accuracy of vehicle dynamics state correction with different Transformer layers.}
    \renewcommand\arraystretch{1.5}
    \setlength{\tabcolsep}{4pt}
	\begin{threeparttable}
    \resizebox{\linewidth}{!}{
		\begin{tabular}{r | r | r r r | r r r }
			\toprule
			\multirow{2}{*}{$D$} & \multirow{2}{*}{param.} &  \multicolumn{3}{c | }{Mean error on \textbf{val1} $\downarrow$ } & \multicolumn{3}{c}{Mean error on \textbf{val2} $\downarrow$}\\
             & & $v_x$ & $v_y$ & $w_z$ & $v_x$ & $v_y$ & $w_z$  \\
			\midrule
            1 & 87.0K & 0.041 & 0.0090 & 0.0019 & 0.14 & 0.0075 & 0.0039 \\
			\rowcolor{gray!40}2 & 170.8K & 0.022 & 0.0044 & 0.00079 & \textbf{0.059} & 0.0062 & \textbf{0.0013} \\
            4 & 338.2K &  \textbf{0.014} & \textbf{0.0032} & 0.00071 & 0.070 & \textbf{0.0051} & 0.0014\\
            6 & 505.6K & 0.036 & \textbf{0.0032} & 0.0010 & 0.15 & 0.010 & 0.0030\\
            8 & 673.0K & 0.030 & \textbf{0.0032} & \textbf{0.00067} & 0.18 & 0.0090 & 0.0024\\
			\bottomrule
		\end{tabular}
    }
    \begin{tablenotes}
        \footnotesize
        \item[] All the DCR methods employ MLPDyTR model.
    \end{tablenotes}
	\end{threeparttable}
	\label{table:depth}
\end{table}

\subsubsection{Effectiveness of Transformer layers:} We investigate the impact of different depth of Transformer layers, $D$, on state correction performance. For simplicity, we set the depth of Transformer encoder layers in the temporal fusion module and Transformer decoder layers in the residual state module to be the same in this work. The results are listed in Tab. \ref{table:depth}. When applying just one single Transformer layer, the model cannot fully aggregate temporal dynamics features and estimate residuals with high precision. The model can achieve excellent performance with 2 or 4 Transformer layers. But its performance degrades with larger $D$. The phenomenon is more significant on \textbf{val2} split, indicating that the model will be struggled with overfitting problem with an excessive depth of Transformer layers. In order to balance efficiency and effectiveness, we choose $D=2$ as the default parameter.

% mode
\begin{table}[t]
	\centering
	\caption{The accuracy of vehicle dynamic state correction with different dynamics residual query.}
    \renewcommand\arraystretch{1.5}
	\begin{threeparttable}
    \resizebox{\linewidth}{!}{
		\begin{tabular}{r | r r | r r r | r r r }
			\toprule
			\multirow{2}{*}{method} & \multicolumn{2}{c|}{query} & \multicolumn{3}{c | }{Mean error on \textbf{val1} $\downarrow$ } & \multicolumn{3}{c}{Mean error on \textbf{val2} $\downarrow$}\\
             & $\hat{\textbf{s}}_{t+1}$ & $c$  & $v_x$ & $v_y$ & $w_z$ & $v_x$ & $v_y$ & $w_z$  \\
			\midrule
			3 DoF & - & - & 0.19 & 0.035 & 0.0080 & 0.13 & 0.015 & 0.0042 \\
            14 DoF & - & - & 0.16 & 0.029 & 0.0081 & 0.12 & 0.014 & 0.0044\\
            \hline
            (a) & \ding{56} & \ding{56} & 0.052 & 0.0084 & 0.0015 & 1.0 & 0.035 & 0.020 \\
            (b) & \ding{56} & \checkmark & 0.031 & 0.0078 & 0.0015 & 0.44 & 0.022 & 0.0093\\
            (c) & \checkmark & \ding{56} & 0.060 & 0.0083 & 0.0020 & 0.069 & 0.0087 & 0.0026\\
            \rowcolor{gray!40}MLPDyTR & \checkmark & \checkmark & \textbf{0.022} & \textbf{0.0044} & \textbf{0.00079} & \textbf{0.059} & \textbf{0.0062} & \textbf{0.0013}\\
			\bottomrule
		\end{tabular}
    }
    \begin{tablenotes}
        \footnotesize
        \item[] All the DCR methods employ MLPDyTR model.
    \end{tablenotes}
	\end{threeparttable}
	\label{table:mode}
\end{table}

\subsubsection{Effectiveness of dynamics residual query:} In the proposed DyTR model, the dynamics residual query $Q$ in the residual estimation module is composed of the prediction from the \textit{base model} $\hat{\textbf{s}}_{t+1}$ and the vehicle configuration $c$. We further investigate the impact of each factor by ablation analysis. The results are listed in Tab. \ref{table:mode}. The variants (a) and (b) discard $\hat{\textbf{s}}_{t+1}$, so they can be regarded as common DNN-based DSE model. Oppositely, the variant (c) and the full MLPDyTR model take $\hat{\textbf{s}}_{t+1}$ as input, so they belong to DRC model in this work. It is clearly shown that the model cannot achieve pleased performance without $\hat{\textbf{s}}_{t+1}$. Particularly, on more challenging \textbf{val2} split, the DNN-based DSE models cannot generalize to various driving conditions and are even worse than physics-based DSE model. The results indicate the superior effectiveness of DRC model compared to DSE model. Compared (a) with (b) and (c) with full MLPDyTR model, the $c$ in the query also promote the generalization ability towards various vehicle configuration, substantiating the rationality of the dynamics residual query $Q$.

\section{Conclusions}
\label{sec:conclusion}

In this letter, we address the complex distributed electric-drive vehicles dynamics modeling problem by proposing a novel Transformer-based DRC network, DyTR, to correct the estimated dynamics state by a simple physics-based model.
The network extracts dynamics feature from historical temporal states and control signals. Then, the dynamics state residual is implicitly represented as high-dimensional features, \textit{i.e.}, dynamics residual query, which is initialized by physics-based model's estimation and vehicle configuration and then iteratively updated by the Transformer-based residual estimation module. By constantly retrieving relevant information from temporal aggregated dynamics features by attention mechanism in a Transformer architecture, the dynamics residual query could regress estimation residuals with high-precision.
To validate the effectiveness of the proposed network, we conduct co-simulation to generate required training and validation data. 
The proposed DyTR network is fully analyzed on the simulation dataset and concluded to be able to significantly reduce the errors of physics-based models by an average of 92.3\% and 59.9\% in two datasets, which is necessary for the requirements of accurate state estimation for high-level autonomous driving.

% \newpage




\vfill

\end{document}